\documentclass[11pt]{article}
\usepackage{coling2020}
\usepackage{times}
\usepackage{url}
\usepackage{latexsym}
\usepackage{graphicx}
\usepackage{booktabs}
\usepackage{todonotes}

\usepackage[switch]{lineno}
\usepackage{enumitem}

\newcommand\figref{Figure~\ref}

\newcommand\tabref{Table~\ref}

\colingfinalcopy 

\title{
    Aspect-Based Argument Mining
}

\author{
    Dietrich Trautmann \\
    Center for Information and Language Processing \\
    Ludwig Maximilian University of Munich, Germany \\ 
    {\tt dietrich@trautmann.me}
}

\date{}

\begin{document}
\maketitle


\begin{abstract}
Computational Argumentation in general and Argument Mining in particular are important research fields.
In previous works, many of the challenges to automatically extract and to some degree reason over natural language arguments were addressed.
The tools to extract argument units are increasingly available and further open problems can be addressed.
In this work, we are presenting the task of Aspect-Based Argument Mining (ABAM), with the essential subtasks of Aspect Term Extraction (ATE) and Nested Segmentation (NS).
At the first instance, we create and release an annotated corpus with aspect information on the token-level.
We consider aspects as the main point(s) argument units are addressing.
This information is important for further downstream tasks such as argument ranking, argument summarization and generation, as well as the search for counter-arguments on the aspect-level.
We present several experiments using state-of-the-art supervised architectures and demonstrate their performance for both of the subtasks.
The annotated benchmark is available at \url{https://github.com/trtm/ABAM}.
\end{abstract}


\blfootnote{
    %
    %
    %
    %
    \hspace{-0.65cm}  
    This work is licensed under a Creative Commons 
    Attribution 4.0 International License.
    License details:\\
    \url{http://creativecommons.org/licenses/by/4.0/}.
}

\section{Introduction}

The field of computational argumentation \cite{slonim2016nlp} gained a lot of interest in the last couple of years.
This is noticeable from both the number of the submitted publications related to this field and also from the high volume of emerging datasets \cite{aharoni2014benchmark,levy2017unsupervised,habernal2018semeval,stab2018cross,trautmann2020fine}, specific task formulations \cite{wachsmuth2017pagerank,al2020end} and models \cite{kuribayashi2019empirical,chakrabarty2020ampersand}.

Similar to aspect-based sentiment analysis \cite{pontiki2014semeval}, we also see the possibility of breaking down arguments into smaller attributes or meaningful components in the argument mining domain.
We consider these components as \emph{aspects} of the arguments.
Previous works already utilized aspect-information for several subtasks within the argument mining domain \cite{fujii2006system,misra2015using,gemechu2019decompositional}.
However, these works vary significantly in the definition of aspects and do not focus on the aspect-based argument mining explicitly, e.g., employ aspects as a source of side or additional information.

For instance, \newcite{fujii2006system} are mainly focusing on the summarization of opinions, visualizing pro and contra arguments for a given topic. 
Thereby, the authors are extracting aspects, calling them \emph{points at issue}, and ranking the arguments according to them.
However, their approach relies on rule-based extraction solely.
In \newcite{misra2015using}, the authors are proposing summarization methods to recognize specific arguments and counter-arguments in social media texts, to further group them across discussions into \emph{facets} (i.e., aspects) on which that issue is argued.
Still, this work is limited to a couple of topics and samples.
Finally, \newcite{gemechu2019decompositional} also mention aspects as part of four functional components, where the authors interchangeably label aspects and concepts for the specific words.
However, to the best of our knowledge, the authors did not publish their labeled data, making a comparative evaluation of aspect extraction methods impossible.
We, in contrast, specifically address the aspect term extraction, concentrate on the proper definition of aspects and therefore directly emphasize and present the task of Aspect-Based Argument Mining ({\small ABAM}) in this work.

\begin{figure}
    \centering
    \includegraphics[width=0.5\textwidth]{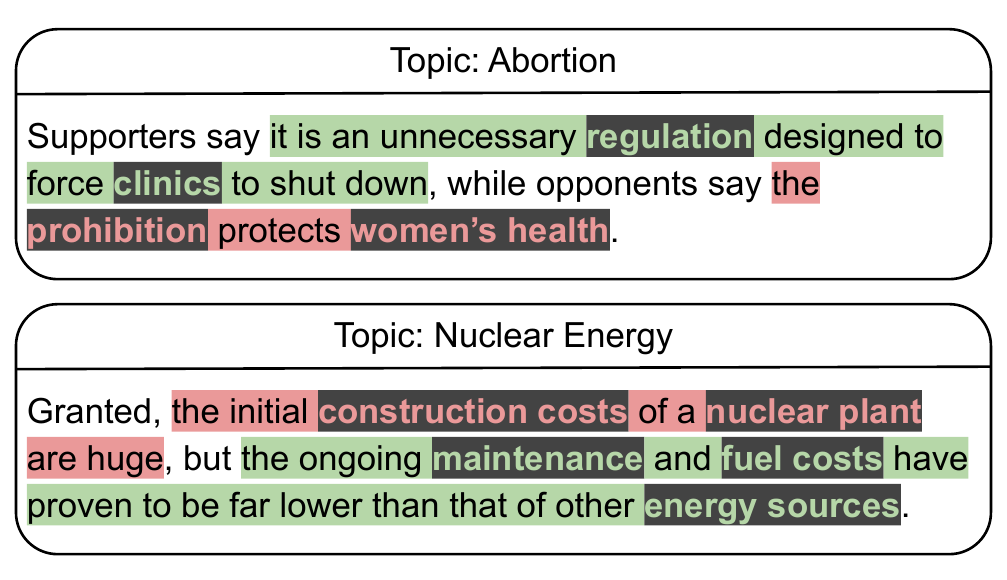}
    \caption{Example annotation of argumentative spans, the corresponding stances
    (green: supporting/pro; red: opposing/contra) and the aspects (underlined) 
    for the topics \emph{abortion} and \emph{nuclear energy}.}
    \label{fig:aspect_term_examples}
\end{figure}

One of the potential applications for the {\small ABAM} is the ability to search for specific subtopics within a larger controversial area.
For instance, for the topic \emph{abortion}, one can particularly be interested in \emph{regulation} or \emph{health}-related aspects (first example in \figref{fig:aspect_term_examples}).
Whereas for the topic of \emph{nuclear energy}, one can care for solely \emph{enviromental}, \emph{cost}- or \emph{safety}-related aspects (second example in \figref{fig:aspect_term_examples}).
By searching or filtering for the particular aspects, one has the possibility to select for specific information and, therefore, to get more fine-grained results.
Another benefit is the ability to compare opposing arguments on the aspect-level.


In this regard, necessary subtasks within the {\small ABAM} include the explicit \emph{Aspect Term Extraction} (ATE) on token-level and the \emph{Nested Segmentation} (NS) of argumentative parts along with their aspects within a given sentence.
Our work is based on \newcite{trautmann2020fine}, where the authors already addressed the task of argument unit segmentation.
We extend their benchmark with aspect term extraction on these argument units.
The {\small ABAM} task can be performed in two ways: 
first, as a two-step pipeline approach with argument unit recognition and classification (AURC) followed by aspect term extraction, 
or as an end-to-end approach in the form of the nested segmentation task. 
Since the argument units are already provided by \newcite{trautmann2020fine}, 
we can use them directly for the second step in the pipeline, namely the ATE task.
Whereas in the end-to-end scenario we adress both tasks (i.e., AURC and ATE) simultaneously for argumentative sentences.

One of the main challenges we faced during this work was the absence of publicly available benchmarks containing the aspect terms.
Existing argument mining datasets do not contain the required information and therefore could not be directly applied for \emph{Aspect-Based} Argument Mining.
We address this challenge by extending an existing fine-grained argument corpus \cite{trautmann2020fine} with crowd-sourced token-level aspect information.
This is our focused main contribution.
While annotating the corpus, we were faced multiple difficulties, including the proper definition of aspects and the creation of rules required for the aspect extraction.
It is important to note, that within this work, we refer to aspects as the main point(s) arguments are addressing. 

Last but not least, since we are extending the existing corpus, we do not explicitly concentrate on the stance definition and its annotation.
Furthermore, as stated in \newcite{trautmann2020fine}, there are two main argument mining directions: \emph{closed domain discourse-level} and the argument mining from the \emph{information seeking} perspective.
The authors of the underlying corpora follow the latter and provide the reasons for that in their work.
We, therefore, adopt their vision on that point. 

Summarizing the abovementioned points, our contribution within this work is as follows:
\begin{itemize}[noitemsep,nolistsep]
    \item We are emphasizing and presenting the task of Aspect-Based Argument Mining on its own.
    \item We are extending an existing corpus with token-level aspect terms, making a comparative evaluation of {\small ABAM} methods possible. 
    \item We are presenting a number of strong baselines with a corresponding error analysis.
\end{itemize}

\section{Problem Statement}
\label{sec:problem}

We define the {\small ABAM} task as following:
Given a list of several topic related texts (documents or paragraphs), we segment the texts into $N$ sentences

\begin{equation}
        sentence_i = [t_1, t_2, t_3, \ldots, t_n] 
\end{equation}

The problem is to select, if available, one (or several) span(s)

\begin{equation}
        span_j = [t_k, \ldots, t_l] 
\end{equation}

inside each $sentence_i$, with $k>=1$, $l<=n$, $l-k>=SEG_{min}$ and $l-k<=SEG_{max}$ 
(with $SEG_{min}=3$ tokens and $SEG_{max}=n$ tokens in a segment), 
and a corresponding stance 

\begin{equation}
        stance_j \in [PRO, CON]
\end{equation}

Tokens outside of argumentative spans are assigned the $NON$ stance label.
Furthermore, regularly there is at least one aspect in every selected span with

\begin{equation}
        aspect_j = [t_p, \ldots,t_q]
\end{equation}

where $p>=k$, $q<=l$, $q-p>=ASP_{min}$ and $q-p<=ASP_{max}$
(with $ASP_{min}=1$ token and $ASP_{max}=5$ tokens per aspect).

\section{Related Work}
\label{sec:rw}
Regarding the abovementioned problem definition (\S\ref{sec:problem}), we selected three research areas as thematically closed to our task.

\paragraph{Sentiment Analysis:}
The SemEval workshop organized the task of aspect-based sentiment analysis
\cite{pontiki2014semeval,pontiki2015semeval,pontiki2016semeval}.
Its subtasks also involved the aspect term extraction, which mainly inspired our approach and definition of the aspect term.
Recent works applied adversarial training of pretrained language models \cite{karimi2020adversarial}
and a combination of contextualized embeddings and hierarchical attention \cite{trusca2020hybrid}
for new state-of-the-art results on this tasks.

\paragraph{Argument Mining:} 
In our work we adopt the definition of argument facets from the previous work and adjust it for our task.
For instance, \newcite{misra2015using} used the information on argument facets for the summarization of arguments in social media.
Furthermore, the authors used argument facets for the argument similarity task \cite{misra2016measuring}.
The abovementioned works were a first approach in the area of argument facet extraction and were limited to solely a couple of topics and samples.
Recent work extended this approach to 28 topics and used the aspect information
for the argument similarity task and argument clustering \cite{reimers2019classification}.
However, the focus of \newcite{reimers2019classification} was on the pairwise classification of argumentative sentences
and not on the aspect term extraction task itself.
Lastly, the work by \newcite{bar2020arguments} defined argument key-points to create 
concise summaries from a large set of arguments.

\paragraph{Nested Named Entity Recognition:} 
The task of nested-NER is similar to the nested segmentation task (\S\ref{task:ns})
that we propose. Early work \cite{finkel2009} presented newspaper and biomedical corpora,
and modeled the data by manual feature extraction.
Recent works proposed recurrent neural networks \cite{katiyar2018nested}
and sequence-to-sequence \cite{strakova2019neural} approaches.
The latter modeled nested labels as multilabels, a method that we also adopted 
for our task with overlapping stance and aspect labels.

\section{Corpus Creation}
\label{sec:cc}
The creation of the {\small ABAM} benchmark is based on the argument units from the AURC corpus \cite{trautmann2020fine} and is divided into two main parts. 
The first part addresses two studies for the annotation task formulation, whereas the second part describes the final corpus creation.
We outsourced the data annotation to independent (crowd-)annotators and based on their results we created the gold labels.

\subsection{Expert Study}
We conducted two expert studies on random samples of ten argument units per stance and topic, selected from the AURC corpus. The resulting sets contained $160$ samples for each study.

\subsubsection{Token-Level Annotation}
\label{sec:tla}
The first expert study task was to select explicit aspect terms from a given argument unit on the token-level.
Two graduate domain experts performed the annotation. Experts were free to select every input-token which fits the following task description:
\emph{``The aspects are defined as the most important point(s) the argument unit is addressing"}.

After the annotation step, the Inter-Annotator Agreement (IAA) for the $160$ samples was computed.
We decided for Cohen's $\kappa$ \cite{cohen1960coefficient} as our agreement measure, that resulted in the initial score of $0.538$.
According to \newcite{viera2005understanding}, this score is in the \emph{moderate agreement} range.
Furthermore, the primary analysis of the selected aspect terms from both annotators yielded a list of especially \emph{frequent part-of-speech} (PoS) patterns for the selected tokens.
To further improve the annotation process, the PoS information was employed in the second expert study.

\begin{table}
\centering
\vspace*{0.2in}
\scriptsize
\scalebox{1.3}{
    \begin{tabular}{r|r|r}
        \toprule
        NN          & NNS JJ NNS     & JJ HYPH NN NN     \\
        NNS         & NN POS NN      & JJ HYPH NN NNS    \\
        NN NN       & NN POS NNS     & JJ HYPH JJ NN     \\
        NN NNS      & NNS POS NN     & JJ HYPH JJ NNS    \\
        JJ NN       & NNS POS NNS    & JJ JJ NN NN       \\
        JJ NNS      & IN NN NN       & JJ JJ NN NNS      \\
        NN NN NN    & IN NN NNS      & JJ NN HYPH NN     \\
        NN NN NNS   & JJ NN NN       & JJ NN HYPH NNS    \\
        NN IN NN    & JJ NN NNS      & JJ NN JJ NN       \\
        NN IN NNS   & JJ JJ NN       & JJ NN JJ NNS      \\
        NN HYPH NN  & JJ JJ NNS      & JJ NN NN NN       \\
        NN HYPH NNS & NN HYPH NN NN  & JJ NN NN NNS      \\
        NN JJ NN    & NN HYPH NN NNS & JJ HYPH NN NN NN  \\
        NN JJ NNS   & NN POS JJ NN   & JJ HYPH NN NN NNS \\
        NNS JJ NN   & NN POS JJ NNS  &                   \\
        \bottomrule
    \end{tabular}
    }
    \caption{The final set of the 44 Part-of-Speech patterns.}
    \label{tab:pos_patterns}
\end{table}

\subsubsection{Candidates Selection}
\label{sec:cs}
The aspect candidate selection step is crucial for the correct aspect term extraction task.
To select the aspect candidates for the second study, we rely on the part-of-speech information.
Specifically, the PoS patterns that occurred more than twice in the previous expert study (i.e., token-level annotation) where picked, and some additional PoS patterns were defined (e.g., the singular and plural form of nouns).
The tag set is based on the Part-of-Speech tags used in the Penn Treebank Project\footnote{\url{https://www.ling.upenn.edu/courses/Fall_2003/ling001/penn_treebank_pos.html}} and the stanza NLP library\footnote{\url{https://stanfordnlp.github.io/stanza/}}.
The final PoS pattern list is comprehensive and representative (includes $44$ patterns, see \tabref{tab:pos_patterns}), and ensures linguistically and grammatically correct candidates, without affecting the actual discourse.
These PoS patterns were applied on a different set of $160$ random samples to create a list of aspect term candidates for every argument unit.

The annotators were asked to solve the same task as before, but now by selecting one or several options from the aspect term candidates list. If none of the aspect term candidates were appropriate, the option \emph{NONE} was selected.
This simplification of the task, compared to the first study, led to a raised Cohen's $\kappa$ of $0.790$. This is considered as a \emph{substantial agreement} \cite{viera2005understanding} and we deem this as a viable approach for the aspect term extraction.

\subsection{Corpus Annotation}

Based on the insights from the first two studies, 
the annotation guidelines (\S\ref{app:annotation_guidelines})
were extended with clearer task formulations and examples.
Additionally, the final set of PoS patterns
(full list in \tabref{tab:pos_patterns}) was applied on all argument units from the AURC corpus.
The AURC corpus was slightly preprocessed to account for duplicates on the sentence- and segment-level, as well as on some minor errors on span boundaries.

Two independent (crowd-)annotators with a linguistic background and a minimum professional working proficiency in English were recruited for the aspect term extraction task.
The annotation procedure was the same as described in \S\ref{sec:cs}.
The inter-annotator agreement score for the two expert annotators resulted in a Cohen's $\kappa$ of $0.874$ for all eight (8) topics.
This is considered as an \emph{almost perfect agreement} \cite{viera2005understanding}.

\paragraph{Annotation Merge}
For the gold standard we selected the annotations where both of the annotators agreed on the token-level.
This ensured that we always had a selection of aspects if neither of the annotators selected the NONE option.
Additionally, shorter aspect terms are favoured by this annotation merge.

\paragraph{Gold Standard}
The final descriptive statistics of the {\small ABAM} corpus are depicted in the \tabref{tab:corpus_stats}.
There are $12040$ aspects in total and $4525$ unique (lemmatized) aspects. 
The topic with the most segments (T8 in \tabref{tab:corpus_stats}), also yielded the most total aspects ($2019$).
Furthermore, there are 58.10\% of the aspects with only one token, 32.12\% with 2 tokens, 7.94\% with 3 tokens, 1.73\% with 4 tokens and only 0.12\% with 5 tokens.


\begin{table*}
\centering
\vspace*{0.2in}
\scriptsize
\scalebox{1.3}{
	\begin{tabular}{llrrrr}
    \toprule
        \textbf{\#} & \textbf{topic} & \textbf{\#sentences} & \textbf{\#segments} & \textbf{\#aspects (total)} & \textbf{\#aspects (unique)} \\
    \midrule
        T1 & abortion                &  415 &  435 &   910 &  484 \\
        T2 & cloning                 &  343 &  365 &   843 &  492 \\
        T3 & marijuana legalization  &  626 &  676 &  1889 &  887 \\
        T4 & minimum wage            &  624 &  689 &  1981 &  745 \\
        T5 & nuclear energy          &  615 &  671 &  1992 &  980 \\
        T6 & death penalty           &  588 &  637 &  1325 &  545 \\
        T7 & gun control             &  480 &  519 &  1081 &  429 \\
        T8 & school uniforms         &  705 &  800 &  2019 &  923 \\
        \midrule
           & total 		             & 4396 & 4792 & 12040 & \textbf{4525}\textsuperscript{\textdagger} \\
    \bottomrule
    \end{tabular}
    }
    \caption{Count of sentences, segments and (total \& unique) aspects in the ABAM corpus. \textsuperscript{\textdagger}Clarification: The total count of unique aspects for all topics is 4525, but the sum of all unique aspects per topic is 5485. This is due to some aspects appearing in several topics (c.f. \tabref{tab:top5}).}
    \label{tab:corpus_stats}
\end{table*}

\paragraph{Common Aspects}
In further aspect analysis we aggregated the most common aspects for the eight topics.
The top five aspects and the absolute occurence counts per topic, are shown in \tabref{tab:top5}.
Furthermore, three aspects (\emph{life, problem, government}) appeared in all eight topics 
and the aspects \emph{people, cost, society, risk, law} appeared in seven topics.

\begin{table}
\centering
\vspace*{0.2in}
\scriptsize
\scalebox{1.2}{
    \begin{tabular}{l|l}
        \toprule
        \textbf{topic}                    & \textbf{aspect (occurrences)} \\
        \midrule
        \midrule
        abortion                          & child (28), life (26), woman (24), unsafe abortion (22), death (16) \\
        \midrule
        cloning                           & animal (24), child (20), clone (20), disease (16), scientist (16) \\
        \midrule
        marijuana legalization            & drug (51), marijuana (44), people (41), alcohol (37), medical marijuana (27) \\
        \midrule
        minimum wage                      & worker (119), job (52), increase (46), employer (41), economy (39) \\
        \midrule
        nuclear energy                    & energy (42), electricity (35), fossil fuel (34), environment (30), nuclear power plant (24) \\
        \midrule
        death penalty                     & crime (62), deterrent (30), punishment (28), cost (27), criminal (27) \\
        \midrule
        gun control                       & crime (56), gun (55), criminal (28), crime rate (25), gun control law (22) \\
        \midrule
        school uniforms                   & student (140), parent (77), child (66), kid (60), school (57) \\
        \midrule
        \midrule
        common aspects (in 8 of 8 topics) & life (91), problem (57), government (55) \\
        \midrule
        common aspects (in 7 of 8 topics) & people (94), cost (78), society (51), risk (48), law (42)\\
        \bottomrule
    \end{tabular}
    }
    \caption{The top 5 most common aspects per topic and for aspects that appear in several topics.}
    \label{tab:top5}
\end{table}

\section{Experimental Setup}
This section presents our experimental setup regarding the two tasks, the employed models and the data set splits.

\subsection{Tasks}
\label{sec:task}
In this work we apply the two different, but related, sub-tasks for {\small ABAM} in the sequence labeling formulation, following \newcite{akhundov2018sequence}.

\subsubsection{Aspect Term Extraction}
\label{task:ate}
In the first task (ATE), we employ only the aspect term information within the segments (argument untis).
This sequence labeling task is a \emph{binary} classification problem per token.

\subsubsection{Nested Segmentation}
\label{task:ns}

In the second task (NS), we utilize full argumentative sentences (like the examples in \figref{fig:aspect_term_examples}) with the 
\emph{stance} (PRO, CON, NON) and \emph{aspect} (O, ASP) information for every token as our input.
We extend the stance labels with the aspect information for a total set of five possible combinations ([NON,O], [PRO,O], [PRO,ASP], [CON,O], [CON,ASP]).\footnote{Tokens that are not part of argument units (spans) get the stance-label NON in this sequence labeling task and aspects are always within argumentative spans.}

This is a \emph{multiclass} sequence labeling problem, which solves both the argument unit segmentation and the aspect term extraction tasks.

\subsection{Models}

\paragraph{BERT}
For the two subtasks, we decided for the BERT model \cite{devlin2019bert} as a 
recent state-of-the-art system on a number of natural language processing tasks. 
We utilize the base and large versions of BERT, as well as both versions of the models with an additional CRF-Layer \cite{sutton2012introduction} as the final classification layer in the architecture.
Further information about hyperparameter search and computing infrastructure are in \S\ref{sec:hyperparameters}, \S\ref{app:hyperparameters} and \S\ref{app:compute}.

\paragraph{PoS Patterns}
Additionally, we applied the PoS-patterns from the aspect candidates creation step we used in \S\ref{sec:cc}.
For the ATE task we labeled all tokens that match the PoS-patterns and report the results as the lower boundary of our approaches.

\subsection{Evaluation}
As the evaluation metric, we report the macro-F1 scores\footnote{\url{https://github.com/chakki-works/seqeval}} for both of our tasks.
Further information about accuracy, precision and recall can be found in \S\ref{app:apr}.

\subsection{Inner-Topic \& Cross-Topic}
For a better understanding of the model performance, we followed the two different dataset splits (domains)
as they were defined for the AURC corpus \cite{trautmann2020fine}.
In the inner-topic split we trained, evaluated and tested our models on the 
same set of topics (T1-T6, \tabref{tab:corpus_stats}).
In the cross-topic split we trained our model on T1-T5, 
selected the best hyperparameter from the evaluation on T6 and tested on T7 and T8.
Detailed sample counts are shown in \tabref{tab:sample_count_ate} and 
\tabref{tab:sample_count_ns} for each task, domain and set.

\begin{table}
\vspace*{0.2in}
    \parbox{.45\linewidth}{
        \centering
        \scriptsize
        \scalebox{1.3}{
            \begin{tabular}{l|r|r}
                \toprule
                \multicolumn{1}{r|}{set \textbackslash domain} & \multicolumn{1}{|c|}{INNER} & \multicolumn{1}{|c}{CROSS} \\
                \midrule
                train                     & 2447                        & 2264 \\
                dev                       &  333                        &  516 \\
                test                      &  693                        & 1319 \\
                \bottomrule
            \end{tabular}
            }
            \caption{Sample counts per set and domain for the aspect term extraction task.}
            \label{tab:sample_count_ate}
    }
    \hfill
    \parbox{.45\linewidth}{
        \centering
        \scriptsize
        \scalebox{1.3}{
            \begin{tabular}{l|r|r}
                \toprule
                \multicolumn{1}{r|}{set \textbackslash domain} & \multicolumn{1}{|c|}{INNER} & \multicolumn{1}{|c}{CROSS} \\
                \midrule
                train                     & 2268                        & 2097 \\
                dev                       &  307                        &  478 \\
                test                      &  636                        & 1185 \\
                \bottomrule
            \end{tabular}
            }
            \caption{Sample counts per set and domain for the nested segmentation task.}
            \label{tab:sample_count_ns}
    }
\end{table}

\section{Results}
This section presents the results for our tasks as described in \S\ref{sec:task}.

\subsection{Tasks}

\subsubsection{Aspect Term Extraction}
The best performing options are the BERT\textsubscript{LARGE} models (\tabref{tab:results_ate}).
Both of them perform similar, but the one with the CRF-layer is slightly better on the development set for inner-topic and the test set for the cross-topic.
The inner-topic scores are higher compared to the more challenging cross-topic set-up, were we evaluate the models on unseen topics.
All the models performed  much better than the lower boundary from the PoS-Patterns Matches.
However, this scores are still bellow the human performance of 0.895.
The human performance on this task is based on the results from the second expert study (\S\ref{sec:cs})

\subsubsection{Nested Segmentation}
The results for NS (\tabref{tab:results_ns}), show that the BERT\textsubscript{LARGE} model outperforms the other listed approaches, except for the development set in the inner-topic set-up.
Furthermore, the cross-topic set-up is also more challenging for this task, compared to the inner-topic setting.

\begin{table}[h!]
\vspace*{0.2in}
    \parbox{.45\linewidth}{
        \centering
        \scriptsize
        \scalebox{1.3}{
            \begin{tabular}{l|c|c|c|c}
                \toprule
                \multicolumn{1}{r|}{domain}                   & \multicolumn{2}{|c|}{INNER}   & \multicolumn{2}{|c}{CROSS} \\
                \midrule
                \multicolumn{1}{r|}{model \textbackslash set} &  dev & test                   &  dev & test \\
                \midrule
                PoS-Patterns Matches                          & .600 & .610 & .518 & .640 \\
                BERT\textsubscript{BASE}                      & .819 & .813 & .673 & .749 \\
                BERT\textsubscript{BASE}+CRF                  & .823 & .812 & .669 & .743 \\
                BERT\textsubscript{LARGE}                     & .830 & \textbf{.821} & \textbf{.683} & .754 \\
                BERT\textsubscript{LARGE}+CRF                 & \textbf{.832} & .818 & .681 & \textbf{.756} \\
                \midrule
                human performance                             & \multicolumn{4}{|c}{.895} \\
                \bottomrule
            \end{tabular}
            }
            \caption{F1 results on the dev and test sets for the inner-topic (INNER) and cross-topic (CROSS) set-ups for the aspect term extraction task. }
            \label{tab:results_ate}
    }
    \hfill
    \parbox{.45\linewidth}{
        \centering
        \scriptsize
        \scalebox{1.3}{
            \begin{tabular}{l|c|c|c|c}
                \toprule
                \multicolumn{1}{r|}{domain}   & \multicolumn{2}{|c|}{INNER} & \multicolumn{2}{|c}{CROSS} \\
                \midrule
                \multicolumn{1}{r|}{model \textbackslash set} &  dev & test                 &  dev & test \\
                \midrule
                BERT\textsubscript{BASE}      & .507 & .465 & .278 & .338 \\
                BERT\textsubscript{BASE}+CRF  & .521 & .480 & .270 & .332 \\
                BERT\textsubscript{LARGE}     & .557 & \textbf{.520} & \textbf{.315} & \textbf{.369} \\
                BERT\textsubscript{LARGE}+CRF & \textbf{.563} & .517 & .293 & .358 \\
                \bottomrule
            \end{tabular}
            }
            \caption{F1 results on the dev and test sets for the inner-topic (INNER) and cross-topic (CROSS) set-ups for the nested segmentation task.}
            \label{tab:results_ns}
    }
\end{table}

\newpage
\subsection{Hyperparameters}
\label{sec:hyperparameters}
For our experimental setup with BERT, we fine-tuned the whole (standard) base and large 
models, as well as both models with an additional final CRF-Layer.
We selected the hyperparameters on the development sets and in particular the learning rate 
(range: $0.00001$ - $0.00009$ in $0.00001$ steps)
and the dropout rate (range: $0$ - $0.5$ in $0.1$ steps).
We used grid search, to cover all possible combinations.
The model parameters were optimized with AdamW \cite{loshchilov2018decoupled}.
The training batch size was 32.
Our reported results are the averages from three runs and one epoch took about 
1 minute for the base models and less than 2 minutes for the large models on average.
We fine-tuned for 10 epochs in the ATE task and for 20 epochs in the NS task.
Detailed numbers of the final hyperparameters for each model and task can be found in the tables in the appendix \S\ref{app:hyperparameters}.

\section{Error Analysis}
Recalling our definition of aspects: They are defined as the main point(s) argument units are addressing.
Furthermore, considering our annotation guidelines in \S\ref{app:annotation_guidelines}, 
the most important point is usually not equal to the given main topic.
An overview of the main errors found during the evaluation of the development sets for the best performing models in the inner- and cross-topic set-ups, is given below.

\paragraph{Aspect Term Extraction}
During the evaluation of ATE results, we observed a number of errors, which we grouped into the following categories:

\begin{itemize}[noitemsep,nolistsep]
    \item The models tend to favour NOUNS in general.
    \item Topic words, such as \emph{abortion} or \emph{marijuana legalization}, are often selected as aspects, which is in conflict with our guidelines.
    \item Phrase constructions like \emph{thread of ...} are often selected as a whole aspect by the models. For the benchmark, we, in contrast, focus on the main representative word of such constructions (e.g., \emph{suicide} vs. \emph{thread of suicide}).
    \item In the case of ADJECTIVE+NOUN, we suggest to avoid general adjectives (e.g. \emph{new} in \emph{new treatments}), whereas focused adjectives that are part of the concept should be selected (e.g. \emph{recreational} in \emph{recreational marijuana}). Our observation is, that models in general could not sufficiently differentiate between such adjectives.
    \item Models lack the understanding of domain-specific phrasems like \emph{in vitro fertilisation} or \emph{life without parole} and tend to select only the nominalized part of them (e.g., \emph{fertilisation}, \emph{parole}).
\end{itemize}

Overall the inner-topic set-up achieved much better performace compared to the cross-topic set-up and both models showed significantly better results over the PoS-Patterns Matches baseline.
However, in the cross-topic set-up we faced more repeated errors, such as the tendency to select topic words as aspects and not sufficient understanding of domain-specific phrasems.

\paragraph{Nested Segmentation}
The typology of the main errors in the NS task is similar to the ATE task. 
Additionally, in the NS task, a number of errors occured due to the wrong assigment of the stance labels, especially in the cross-topic set-up.
These results confirm the insight from \newcite{trautmann2020fine}, where most of the errors arose due to the wrong stance classification.
Apparently, the BERT-based models tend to attach to sentiment words for the stance predictions, which is not always correlated.

\section{Conclusion}
{\small ABAM} is a challenging task that, to the best of our knowledge, was not directly addressed before.
We made two important contributions:
First, we created and released a publicly available benchmark for Aspect-Based Argument Mining.
Second, we showcased several baselines for the two subtasks, namely the Aspect Term Extraction and the Nested Segmentation, and performed an elaborative error analysis.
We believe that these findings as well as the benchmark are of high potential for further downstream tasks, such as argument ranking, argument summarization and the search for counter-arguments on the aspect-level.

For the future work, we foresee the investigation of unsupervised approaches for the Aspect Term Extraction task, since they showed promising results within the Aspect-Based Sentiment Analysis domain.
Furthermore, it would be of high interest to incorporate topic-specific knowledge (e.g., understanding of phrasems) into the models to address the discussed error types.
In another line of work, one could also explore distant supervision \cite{rakhmetullina2018distant} or domain adaptation methods \cite{marz2019domain}, as well as relational approaches \cite{trautmann2020relational} for this task.


\bibliographystyle{coling}
\bibliography{abam}


\appendix

\section{Annotation Guidelines}
\label{app:annotation_guidelines}

Annotation guidelines defined for the Aspect Term Extraction task in Aspect-Based Argument Mining.

\paragraph{Task Description}
\begin{itemize}
    \item Given a main topic and an argumentative segment (unit), please select one or several options from the aspect candidates list. 
    \item If no aspect candidate could be selected from the list, pick the option \emph{None}.
\end{itemize}

While selecting the aspects, please consider the following rules:
\begin{itemize}
    \item An aspect is defined as the most important/relevant point for the argument made.
    \item The most important point is usually not equal to the given main topic.
    \item In case of doubt, shorter aspects candidates (generic terms; e.g. ``life span") are prefered over longer candidates (e.g. ``prolonged life span").
\end{itemize}

\paragraph{General Hints}
\begin{itemize}
    \item The selected aspect(s) should be related to the topic in general.
    \item The presence of AND/OR (usually) denote multiple aspects:
    \begin{itemize}
        \item If a sentence contains multiple phrases (e.g., ``abortion causes breast cancer AND it kills unborn children.");
        \item If there is an enumeration and objects connected by AND/OR (e.g. ``abortion causes breast cancer, infertility and pain.");
    \end{itemize}
    \item In the case of ADJECTIVE+NOUN, general adjectives should be avoided (e.g. ``new" in ``new treatments"), whereas focused adjectives that are part of the concept should be selected (e.g. ``recreational" in ``recreational marijuana").
    \item Please, use these test-questions for yourself while annotating: 
    \begin{itemize}
        \item Do you want this argument to be shown to someone, if they select this aspect(s) of the topic, or are other aspect terms in this argument more relevant for the point made?
        \item Which words make you understand the argument most?
        \item Which words are the most relevant and mainly form the meaning of the argument made?
        \item If you would compress the argument into a few most relevant words, which words would that be?
    \end{itemize}
\end{itemize}

\section{Hyperparameters}
\label{app:hyperparameters}
The dropout rate of $0.1$ was always the best option.
The learning rates for the different models are displayed in \tabref{tab:ate_hyperparameters}
for the ATE task and in \tabref{tab:ns_hyperparameters} for the NS task.

\begin{table}
\vspace*{0.2in}
    \parbox{.45\linewidth}{
        \centering
        \scriptsize
        \scalebox{1.3}{
            \begin{tabular}{l|c|c}
                \toprule
                \multicolumn{1}{r|}{domain}   & \multicolumn{1}{|c|}{INNER} & \multicolumn{1}{|c}{CROSS} \\
                \midrule
                BERT\textsubscript{BASE}      & $6e-5$ & $8e-5$ \\
                BERT\textsubscript{BASE}+CRF  & $9e-5$ & $9e-5$ \\
                BERT\textsubscript{LARGE}     & $9e-5$ & $9e-5$ \\
                BERT\textsubscript{LARGE}+CRF & $9e-5$ & $8e-5$ \\
                \bottomrule
            \end{tabular}
            }
            \caption{Hyperparameters (learning rate) for the ATE task.}
            \label{tab:ate_hyperparameters}
    }
    \hfill
    \parbox{.45\linewidth}{
        \centering
        \scriptsize
        \scalebox{1.3}{
            \begin{tabular}{l|c|c}
                \toprule
                \multicolumn{1}{r|}{domain}   & \multicolumn{1}{|c|}{INNER} & \multicolumn{1}{|c}{CROSS} \\
                \midrule
                BERT\textsubscript{BASE}      & $7e-5$ & $5e-5$ \\
                BERT\textsubscript{BASE}+CRF  & $8e-5$ & $6e-5$ \\
                BERT\textsubscript{LARGE}     & $5e-5$ & $7e-5$ \\
                BERT\textsubscript{LARGE}+CRF & $7e-5$ & $8e-5$ \\
                \bottomrule
            \end{tabular}
            }
            \caption{Hyperparameters (learning rate) for the NS task.}
            \label{tab:ns_hyperparameters}
    }
\end{table}

\section{Compute Resources}
\label{app:compute}
We used Kaggle's Kernels\footnote{\url{https://www.kaggle.com/kernels}} for the processing of the data
and Google's Colab\footnote{\url{https://colab.research.google.com/signup}} for the training (fine-tuning) of our models.
The former service offers a single 12GB NVIDIA Tesla K80 GPU, while the latter a single 16GB NVIDIA Tesla P100 GPU.

\section{Additional Results}
\label{app:apr}
The additionally reported numbers for accuracy, precision and recall can be found in the
\tabref{tab:additional_results_ate} for the ATE task, in the 
\tabref{tab:additional_results_ns_args} for the NS task.
The numbers are the average from three runs.


\begin{table*}[h!]\centering
\scriptsize
\scalebox{1.3}{
    \begin{tabular}{l|c|c|c|c|c|c|c|c|c|c|c|c}
        \toprule
        \multicolumn{1}{r|}{domain}   & \multicolumn{6}{|c|}{INNER} & \multicolumn{6}{|c}{CROSS} \\
        \midrule
        \multicolumn{1}{r|}{set}      & \multicolumn{3}{|c|}{dev} & \multicolumn{3}{|c|}{test} & \multicolumn{3}{|c|}{dev} & \multicolumn{3}{|c}{test} \\
        \midrule
        \multicolumn{1}{r|}{model \textbackslash metric} & acc. & pre. & rec. & acc. & pre. & rec. & acc. & pre. & rec. & acc. & pre. & rec. \\
        \midrule
        PoS-Patterns Matches          & .850 & .490 & .773 & .853 & .502 & .779 & .825 & .404 & .724 & .870 & .530 & .809 \\
        BERT\textsubscript{BASE}      & .943 & .784 & .858 & .942 & .786 & .842 & .878 & .580 & .804 & .912 & .682 & .830  \\
        BERT\textsubscript{BASE}+CRF  & .945 & .789 & .860 & .942 & .789 & .836 & .877 & .575 & .800 & .911 & .678 & .822  \\
        BERT\textsubscript{LARGE}     & .946 & .799 & .864 & .945 & .803 & .840 & .881 & .582 & .827 & .914 & .686 & .835  \\
        BERT\textsubscript{LARGE}+CRF & .948 & .798 & .869 & .943 & .802 & .835 & .880 & .585 & .817 & .914 & .690 & .837  \\
        \bottomrule
    \end{tabular}
    }
    \caption{Accuracy (acc.), precision (pre.) and recall (rec.) results on the dev and test sets for the inner-topic (INNER) and cross-topic (CROSS) set-ups for the aspect term extraction task. These are the average scores from three runs.}
    \label{tab:additional_results_ate}
\end{table*}


\begin{table*}[h!]\centering
\scriptsize
\scalebox{1.3}{
    \begin{tabular}{l|c|c|c|c|c|c|c|c|c|c|c|c}
        \toprule
        \multicolumn{1}{r|}{domain}   & \multicolumn{6}{|c|}{INNER} & \multicolumn{6}{|c}{CROSS} \\
        \midrule
        \multicolumn{1}{r|}{set}      & \multicolumn{3}{|c|}{dev} & \multicolumn{3}{|c|}{test} & \multicolumn{3}{|c|}{dev} & \multicolumn{3}{|c}{test} \\
        \midrule
        \multicolumn{1}{r|}{model \textbackslash metric} & acc. & pre. & rec. & acc. & pre. & rec. & acc. & pre. & rec. & acc. & pre. & rec. \\
        \midrule
        BERT\textsubscript{BASE}      & .704 & .468 & .552 & .672 & .434 & .501 & .560 & .234 & .343 & .574 & .296 & .395  \\
        BERT\textsubscript{BASE}+CRF  & .710 & .482 & .568 & .683 & .450 & .515 & .553 & .231 & .324 & .571 & .298 & .376  \\
        BERT\textsubscript{LARGE}     & .748 & .512 & .610 & .709 & .491 & .552 & .597 & .266 & .385 & .607 & .327 & .423  \\
        BERT\textsubscript{LARGE}+CRF & .749 & .524 & .608 & .702 & .491 & .547 & .575 & .248 & .358 & .594 & .320 & .407  \\
        \bottomrule
    \end{tabular}
    }
    \caption{Accuracy (acc.), precision (pre.) and recall (rec.) results on the dev and test sets for the inner-topic (INNER) and cross-topic (CROSS) set-ups for the nested segmentation task (args). These are the average scores from three runs.}
    \label{tab:additional_results_ns_args}
\end{table*}


\end{document}